\documentclass[conference]{IEEEtran}
\IEEEoverridecommandlockouts
\usepackage{cite}
\usepackage{amsmath,amssymb,amsfonts}
\usepackage{algorithmic}
\usepackage{graphicx}
\usepackage{textcomp}
\usepackage{xcolor}
\usepackage{tabularx}
\usepackage{xurl}
\usepackage{todonotes}
\usepackage[hidelinks,breaklinks=true]{hyperref}
\def\BibTeX{{\rm B\kern-.05em{\sc i\kern-.025em b}\kern-.08em
    T\kern-.1667em\lower.7ex\hbox{E}\kern-.125emX}}
\begin{document}

\title{CORNAV: Construction-Aware Reasoning for Robot Navigation on Active Worksites\\
}

\author{Parastoo Ali Pour$^{1}$, Deepak Prakash Kumar$^{2}$, Tommy Zhou$^{3}$, Pramod Khargonekar$^{3}$, Mohammad Abdullah Al Faruque$^{3}$% <-this % stops a space
\thanks{This work was supported in part by Genova Construction.}% <-this % stops a space
\thanks{$^{1}$ The author is with the Department of Mechanical and Aerospace Engineering, University of California at Irvine, Irvine, CA 92697 USA
        {\tt\small alipourp@uci.edu}}%
\thanks{$^{2}$The author is with the Department of Aerospace Engineering, Indian Institute of Technology Bombay, Mumbai 400076, India
        {\tt\small deepakpk@aero.iitb.ac.in}}%
\thanks{$^{3}$The authors are with the Department of Electrical Engineering and Computer Science, University of California at Irvine, Irvine, CA 92697 USA
        {\tt\small \{tommyz8, pramod.khargonekar, alfaruqu\}@uci.edu}}%
}

\maketitle

\begin{abstract}

The construction industry faces persistent labor shortages, low productivity that costs the global economy over \$1.6 trillion annually, and one of the highest injury rates among major industries. These factors motivate the use of autonomous robots to improve efficiency and worker safety. Existing language-grounded navigation systems, however, rely on semantic scene understanding alone and lack access to construction-specific context such as architectural plans, evolving work schedules, and safety constraints. As a result, they localize permanent building features unreliably and cannot safely navigate active jobsites. We present CORNAV, a blueprint-grounded, schedule-aware navigation framework that operates from 2D CAD drawings and project schedules without requiring a Building Information Model. CORNAV aligns architectural blueprints against hierarchical open-vocabulary 3D scene graphs to ground object queries, converts project schedules into time-varying navigation constraints, and validates requests through an LLM-based safety module that escalates hazardous zones before planning. An A* planner then enforces mandatory exclusion zones while preferentially avoiding higher-risk areas. Across an indoor office and a real construction site, blueprint grounding raises task success from 13.0\% to 72.2\% over semantic retrieval alone, schedule awareness eliminates all hard-zone violations, and the safety module correctly rejects hazardous requests arising from mislabeled project schedules. Project website: \url{https://cornav.github.io/CORNAV/}. 
\end{abstract}

\section{Introduction}

The construction industry continues to face persistent workforce shortages \cite{Uthai2025}. In the United States alone, more than 80\% of contractors report shortages of experienced skilled workers \cite{agc2024outlook}, and two-thirds of firms are actively seeking to hire despite ongoing recruitment challenges \cite{agc2024surveyresults}. Construction remains one of the most hazardous industries, with more than 1,000 worker fatalities reported annually since 2016 \cite{cpwr2026fatalitymap}. This industry has also experienced a prolonged decline in productivity relative to other sectors \cite{Uthai2025, goolsbee2023construction}. In the United States, construction value added per worker was roughly 40\% lower in 2020 than in 1970, while manufacturing labor productivity increased more than nine-fold over the same period \cite{goolsbee2023construction}. The economic cost of this gap is substantial and closing it could raise the sector's value added by roughly \$1.6 trillion per year globally, about one-third of that in the United States \cite{mckinseyglobalinstitute}. 

\begin{figure}[tb!]
    \centering
    \includegraphics[width=1\linewidth]{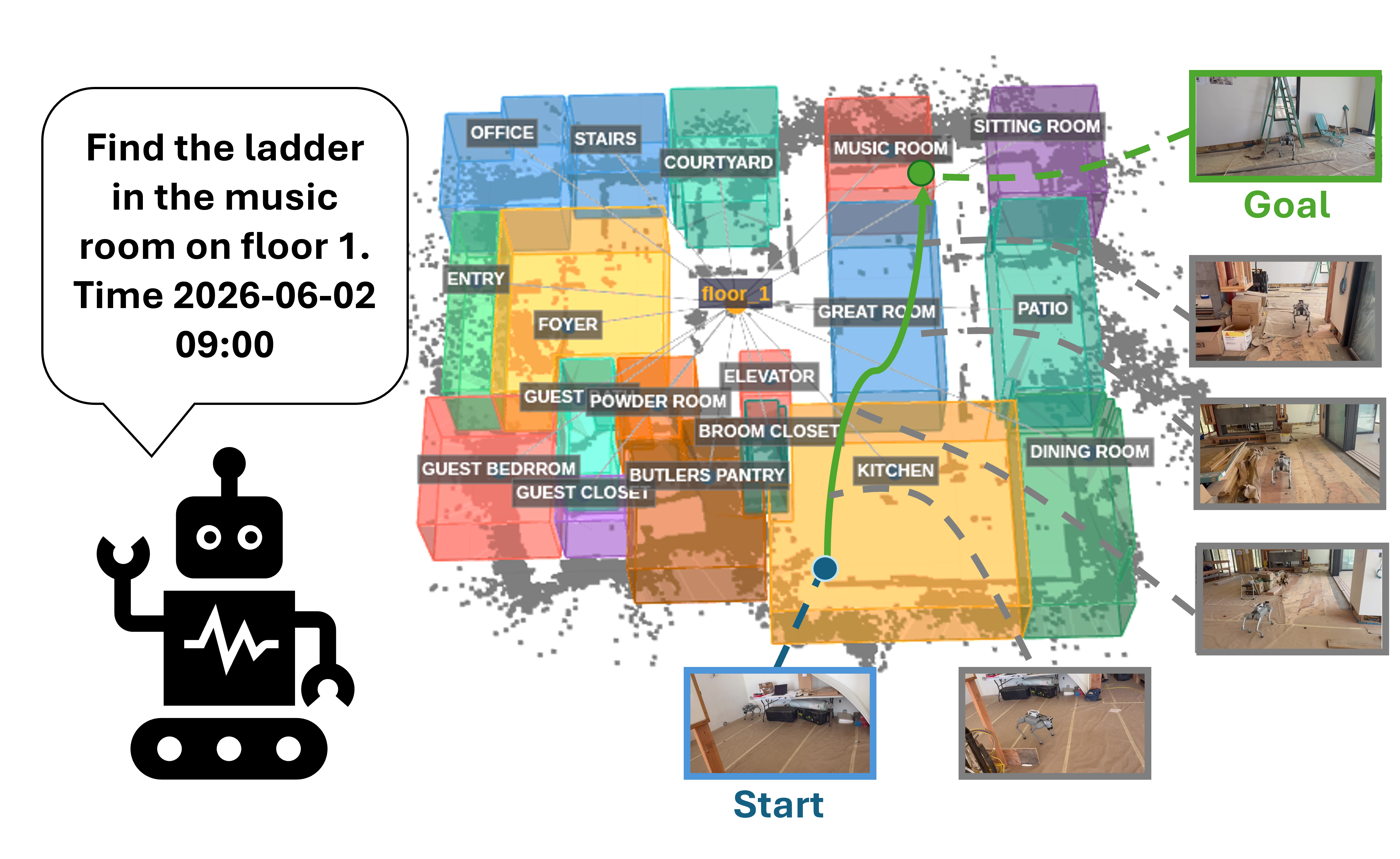}

    \caption{CORNAV framework integrates open-vocabulary scene understanding with construction-specific knowledge including blueprints, project schedules, and safety constraints, to ground natural-language instructions and generate semantically informed, schedule-compliant, safety-validated navigation plans.}
    \label{Fig 1.}
\end{figure}

These challenges, together with the growing need to build and renew aging infrastructure, have accelerated interest in construction robotics. Applications include site inspection and quality control, material handling and transport, assembly and installation, and operation in hazardous environments \cite{Uthai2025}. By automating repetitive and dangerous tasks, robotic systems can improve worker safety, increase operational efficiency, and support more consistent project execution.

Construction sites present a demanding environment for autonomous robots. Unlike structured manufacturing settings, where robots operate within fixed work cells and standardized workflows, construction environments change continuously. Partially completed structures, dynamic equipment, and shifting work zones alter the site as the project progresses \cite{MELENBRINK2020103312, LI2025106005}. Activities such as concrete placement, material staging, and structural installation can change traversable regions and render previously generated maps outdated, making robust localization and frequent map updates essential. To protect workers, construction managers routinely establish temporary work zones, exclusion areas, and task-specific safety procedures that may change throughout the day \cite{Uthai2025}. Autonomous robots must, therefore, reason beyond geometric obstacle avoidance, incorporating semantic understanding, operational context, and evolving safety constraints to navigate active jobsites safely.

Recent work addresses these requirements independently: (i) 3D scene graphs organize objects and spaces into structured semantic representations \cite{armeni20193d, hughes2022hydrarealtimespatialperception}, (ii) open-vocabulary methods add natural-language querying over them \cite{hovsg, gu2024conceptgraphs}, (iii) architectural floor plans and Building Information Models (BIMs) supply geometric and semantic priors for localization and mapping \cite{boniardi2017robust, wang2019glfp, shaheer2023robot}, and (iv) large language models (LLMs) enable task planning and safety reasoning \cite{yang2024safetychip}.

These capabilities, however, have developed largely in isolation. First, scene graph methods offer rich semantic understanding but derive room-level structure from geometric heuristics or vision-language labeling, which varies across scenes and sensing conditions \cite{hughes2022hydrarealtimespatialperception, hovsg, gu2024conceptgraphs}. Second, blueprint- and BIM-based approaches supply stable geometric priors \cite{boniardi2017robust, wang2019glfp, shaheer2023robot} but, as commonly deployed, keep schedule information separate from the geometric model and treat the workspace as static. Third, safety-aware planning methods enforce constraints specified by the robot designer rather than incorporating constraints from the operational documents construction managers already maintain \cite{yang2024safetychip}. Robots therefore lack a unified representation of what exists in the environment, where it is located, when hazards are active, and which safety constraints apply.

\textbf{In summary, this work makes the following contributions.} \textbf{\emph{First}}, we present a unified construction-aware navigation framework that integrates architectural CAD drawings, open-vocabulary 3D scene graphs, construction schedules, and predefined safety rules into a single planning pipeline for autonomous mobile robots (Fig.~\ref{Fig 1.}). \textbf{\emph{Second}}, we introduce a hybrid semantic grounding strategy that combines object retrieval \cite{hovsg} with deterministic blueprint-based room resolution. This enables reasoning over observed objects and architectural fixtures in the CAD while maintaining stable room identities for downstream schedule reasoning. \textbf{\emph{Third}}, we incorporate construction look-ahead project schedules (refer to the project page for an example) directly into navigation. An LLM interprets hazard descriptions and site safety rules to augment hazard labels specified in the project schedule, allowing constraints to adapt as work progresses. \textbf{\emph{Finally}}, we evaluate the framework on robot-collected sensor data (RGB, depth, and pose) from a Unitree Go2 quadruped in an indoor environment and an active construction site, and demonstrate live deployment on both the Go2 and a Unitree G1 humanoid. 

\begin{figure*}
    \centering
    \includegraphics[width=0.75\textwidth]{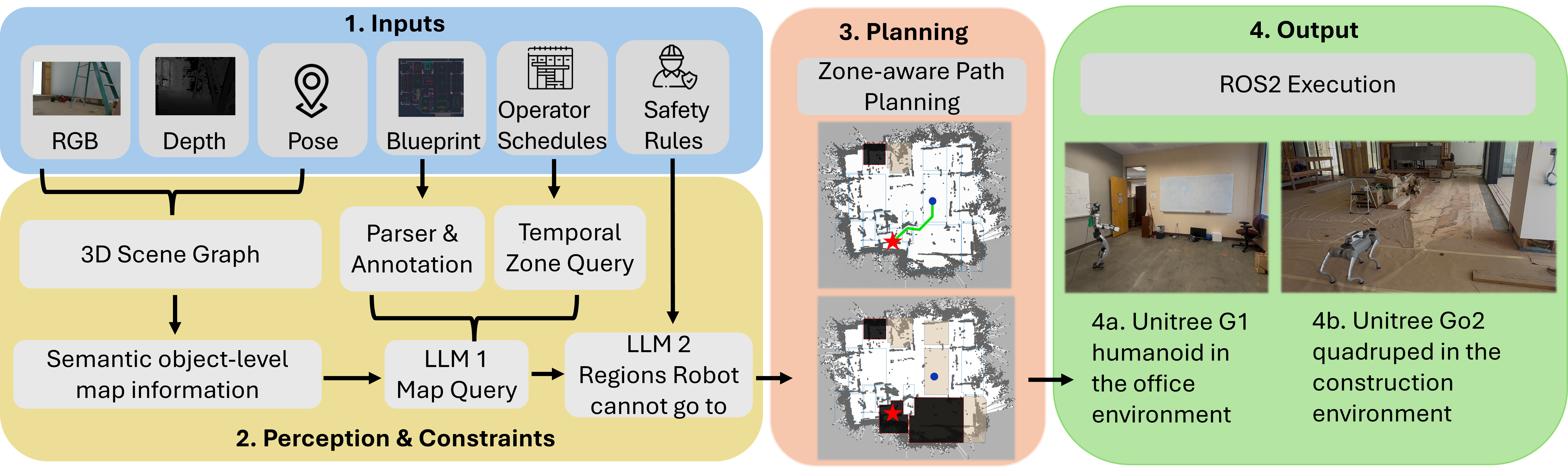}

    \caption{Pipeline of the proposed system. Offline pre-processing runs once per site; the four-stage online pipeline runs per query, taking a natural-language task (e.g., ``find the chair in office 1 on floor 0'') and a requested date and time, returning either robot-frame waypoints or an infeasibility verdict with a reason. In the planning maps, the blue circle marks the robot start, the red star the goal, and green the planned path.}
    \label{Fig 2.}
\end{figure*}

\section{Related Work}

\subsection{Architectural Drawings for Robot Localization}
Architectural drawings have long supplied geometric priors for robot localization and navigation. A range of methods localize against them, using LiDAR \cite{boniardi2017robust, wang2019glfp}, RGB-D imagery, or learned room-layout features \cite{boniardi2019room} to improve robustness in indoor environments with relatively stable layouts. Recent work extends architectural plans to construction-site localization through situational graphs and semantic building representations \cite{shaheer2023robot}. Collectively, these approaches use architectural information primarily for spatial localization and mapping, without accounting for the temporal and safety constraints introduced by ongoing construction activities.

We use architectural drawings differently. Rather than treating the blueprint solely as a localization prior, we use CAD drawings as the authoritative spatial representation for room identities, spatial hierarchy, and safety constraint grounding. This yields stable room assignments robust to errors in perception-based room classification, and provides the backbone for integrating project schedules and safety information.

\subsection{3D Scene Graphs for Semantic Environment}

Recent advances in 3D scene understanding allow robots to build semantic representations of the environment. Hydra introduced real-time hierarchical 3D scene graphs for spatial structure and semantic entities \cite{hughes2022hydrarealtimespatialperception}. Building on advances in open-vocabulary 3D perception \cite{lu2023ovir}, open-vocabulary scene graphs such as OVSG and ConceptGraphs enable language-based grounding and retrieval of objects and regions \cite{chang2023context, gu2024conceptgraphs}. HOV-SG organizes them hierarchically across floors, rooms, and objects for language-grounded navigation \cite{hovsg}. These representations offer far richer environmental understanding than occupancy maps.

Our framework builds the open-vocabulary 3D scene graph using HOV-SG \cite{hovsg}, while recognizing that such graphs alone are insufficient for construction environments, as they do not reliably ground room identity. A region such as a kitchen or living area may be mislabeled in the scene graph, making it hard to associate operational knowledge, such as scheduled hazardous work zones, with the correct region. We therefore combine 3D scene graphs with blueprint-derived spatial information and construction-specific operational knowledge. 
\subsection{BIMs for Robot Navigation} 

BIMs support robotic inspection, localization, and task planning by providing detailed semantic representations of buildings \cite{shaheer2023robot}, and can be linked to project schedules through 4D BIM workflows. However, they require significant modeling effort and ongoing maintenance \cite{REHMAN2025107091}, and adoption remains slow and uneven \cite{vidalakis2019bim}, particularly among the small and medium-sized firms that make up most of the industry \cite{ec_construction, cpwr2026smallfirms}. Limited budgets, training costs, and the perceived adequacy of established 2D architectural drawings sustain the reliance on traditional CAD drawings \cite{li2019critical, vidalakis2019bim}. 4D BIM in particular sees limited uptake \cite{swallow2019benefits}, and schedule information is typically maintained separately from the geometric model.

Closest to our setting, several systems generate schedule-aware navigation maps from BIM. Follini \textit{et al.} \cite{robotics10010002} incorporate construction schedules into Industry Foundation Classes (IFC) geometry to produce a schedule-aware obstacle map, and Gopee \textit{et al.} \cite{gopee2023ifc} block spaces containing active work while generating schedule-dependent waypoints. Others use BIM semantics for routing and task planning. Karimi \textit{et al.} \cite{karimi2021semanticnavigationusingbuilding} build an IFC-derived hypergraph where room attributes (such as wall material 
and construction hazards) determine per-room costs, and routing minimizes total path cost. Kim and Peavy \cite{KIM2022104247} convert IFC into a queryable building world for element-level task planning. In parallel, recent work combines BIM with natural-language risk cues to shape navigation behavior. Amani \textit{et al.} \cite{amani2025bayesian} use Bayesian inference over building data and prompt sentiment to adjust potential fields during planning, biasing the robot away from obstacles when a command signals elevated risk.

These systems demonstrate the value of schedule- and safety-aware navigation, but each presupposes a BIM or IFC model and treats declared hazards as static exclusion zones. They also resolve queried entities through structured lookup over model classes rather than open-vocabulary grounding, so they cannot interpret instructions referring to objects observed on site but absent from the model. Methods that infer risk from prompt tone \cite{amani2025bayesian} face a further limitation. They adjust behavior from the phrasing of a command rather than validating declared hazards against predefined safety rules.

Our framework addresses these gaps in the following ways. We draw operational knowledge directly from the 2D CAD drawings and project schedules already maintained on site, rather than from a BIM or IFC model. Instead of treating scheduled zones as static obstacles, we distinguish mandatory exclusion zones (hard constraints) from cautionary regions (soft constraints). We use an LLM-based validation module to escalate hazardous activity zones to hard constraints, guarding against mislabeled annotations in the project schedules. Finally, an open-vocabulary 3D scene graph lets us ground natural-language queries to both blueprint-defined fixtures and movable objects observed on site, extending beyond the structured lookups of BIM-based systems.

\section{System Overview}

As shown in Fig.~\ref{Fig 2.}, the framework integrates architectural blueprints, semantic scene understanding, construction schedules, and safety rules into a unified navigation pipeline. Offline pre-processing builds a semantic world model and construction knowledge representation from engineering documents; online execution combines these with user task requests to produce schedule-aware, safety-compliant navigation plans.

\subsection{Offline Pre-Processing}

\begin{figure}
    \centering
    \includegraphics[width=\linewidth]{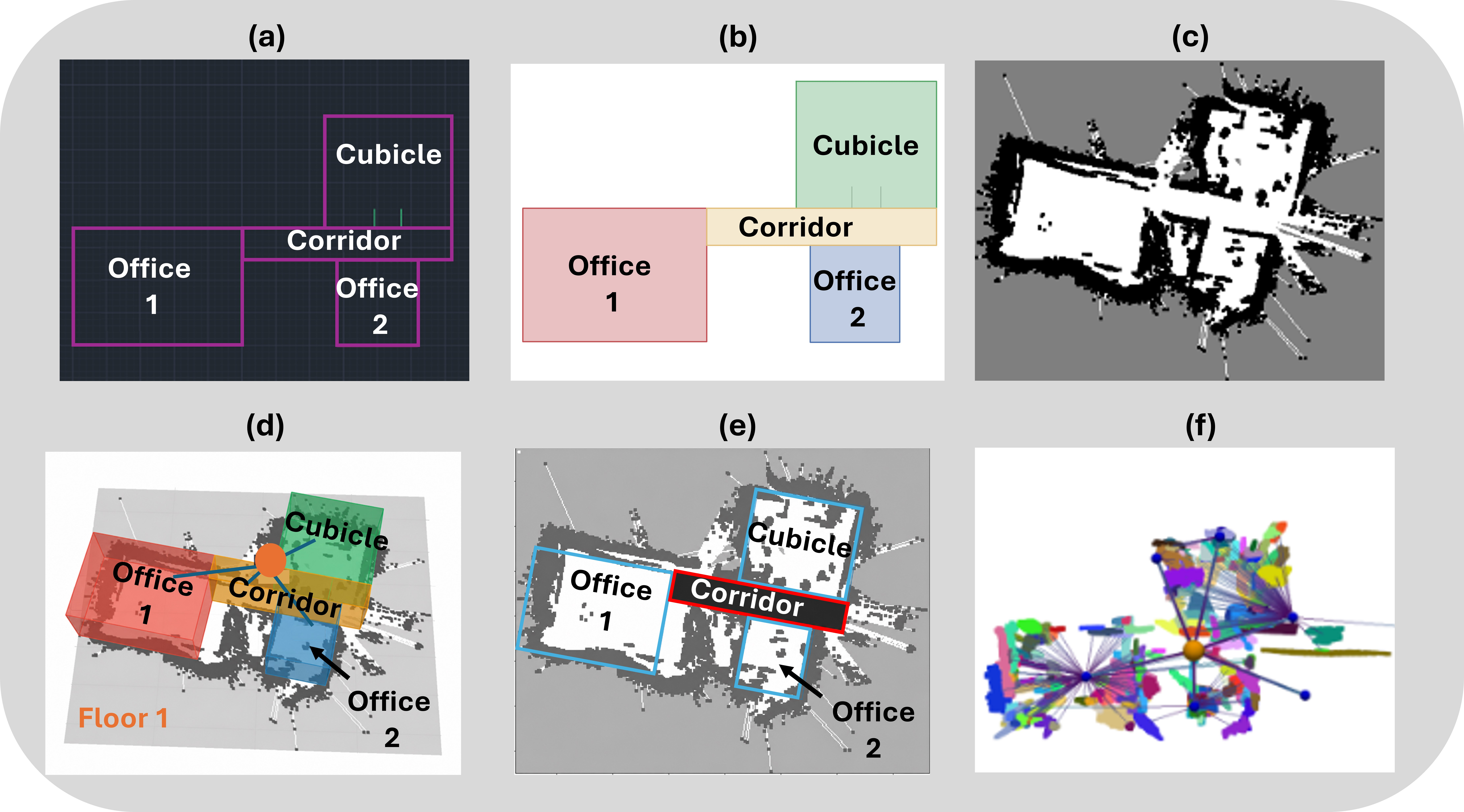}
    \caption{Offline map pre-processing. (a) native CAD drawing,
    (b) extracted room boundaries, (c) robot-generated occupancy map,
    (d) blueprint scene graph aligned to the occupancy map,
    (e) sample project schedule applied to the occupancy map, with
    free space in white and a hard exclusion zone in black, and
    (f) 3D scene graph built from robot observations \cite{hovsg}.}
    \label{fig:map}
\end{figure}

\begin{figure}
    \centering
    \includegraphics[width=0.8\linewidth]{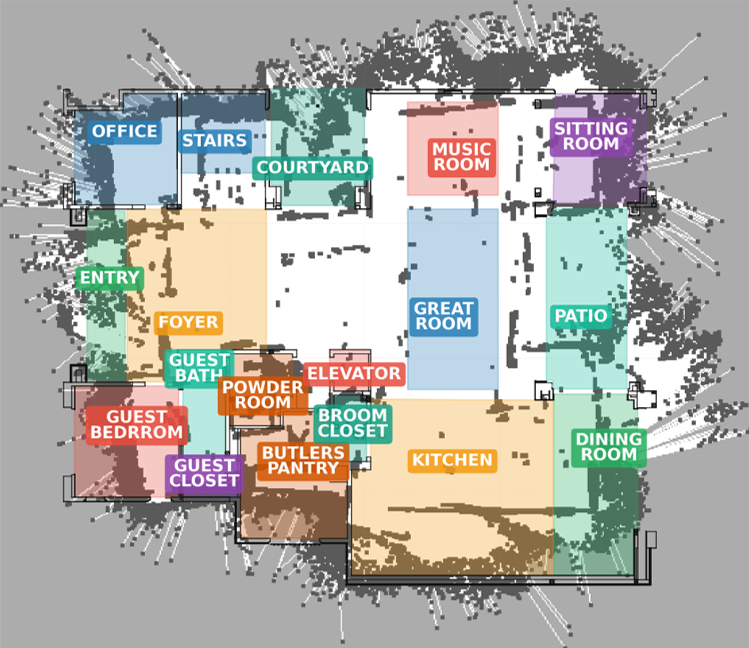}
    \caption{Blueprint overlaid on the occupancy map after alignment, floor 1.}
    \label{Fig 3.}
\end{figure}

Prior to deployment, three sources of information are processed offline (Fig.~\ref{fig:map}). First, the architectural floor plan, exported from CAD to the Drawing Exchange Format (DXF), is parsed to extract room boundaries encoded as closed polylines. These are assembled into a blueprint scene graph containing room polygons, centroids, floor assignments, and spatial hierarchy. The blueprint frame is then aligned to the robot-generated occupancy map by two-stage rigid alignment. A coarse rotational search over rasterized walls provides an initial estimate, refined by iterative closest point (ICP) against the walls the robot observes. Fig. ~\ref{Fig 3.} shows this for one floor (construction floor 1).

Second, project schedules maintained as spreadsheets are converted into a structured JSON representation containing construction activities, room assignments, active time windows, and hazard severity. Moreover, hazard notes are available, which describe potential hazards in each area. This supplies the temporal information needed to activate and deactivate navigation constraints during planning.
Finally, RGB-D observations from an initial site survey are processed with HOV-SG \cite{hovsg} to construct an open-vocabulary 3D scene graph containing semantic object labels, spatial relationships, and CLIP embeddings for language-based retrieval.

\subsection{Object Grounding and Room Resolution}

During execution, natural-language task requests are grounded using complementary information from the blueprint and the 3D scene graph. Transient objects observed during the site survey are retrieved from the HOV-SG representation by CLIP-based semantic matching \cite{hovsg}, via cosine similarity between the query's text embedding and each candidate's visual embedding. Permanent architectural fixtures specified in the CAD drawing (e.g., toilets), by contrast, are resolved directly from the blueprint's CAD-derived registry, bypassing CLIP entirely. This lets the framework reason about both the current state of the environment and planned features not yet present or visible during mapping. Once identified, the object's associated room is resolved deterministically by checking whether its coordinate falls within the queried room's aligned blueprint boundary. 

\subsection{Temporal Constraint}

Construction schedules are queried at the requested execution time to determine which work activities are active. Each activity specifies a room assignment, activity description, hazard notes, and an assigned hazard severity (hard or soft). The corresponding room polygon is retrieved from the aligned blueprint and transformed into the robot frame, producing a set of active geometric constraints. Hard constraints are mandatory exclusion zones where robot entry is prohibited (e.g., active hazardous work or wet concrete). Soft constraints are cautionary regions that remain traversable but are preferentially avoided, unless elevated to hard constraints by the safety validation module.

\subsection{Safety Validation}

Active hard constraints are accepted as mandatory exclusion zones without further reasoning. Active soft constraints are evaluated by a rule-guided language model (GPT-4o) using the scheduled activity notes. The LLM system prompt embeds the project's construction safety rules and hazard keywords drawn from each rule (e.g., scaffolding, unguarded edges). The user prompt contains the navigation query, target object, room, and floor, together with the hazard notes for zones marked active at the query time. The model scores each active soft zone against the embedded safety rules and returns a hazard score from 0 to 1 per zone. Zones scoring at or above a threshold of 0.7 (a design constant) are promoted to hard constraints before planning. Scoring runs at zero temperature for reproducible escalation decisions, after which the validated constraint set is then passed to the navigation planner. The module evaluates active soft zones against the following rules derived from U.S. construction safety regulations (OSHA 29 CFR 1926) \cite{osha1926}:
   
    \begin{itemize}
        \item SR1: No traversal of unapproved temporary structures, including scaffolding, temporary decks, loose formwork, ladders, or any surface not explicitly designed for pedestrian load.
        \item SR2: Avoid unprotected edges and fall hazards. Open slabs, stair openings, balconies without guardrails, and elevator shafts are treated as hard exclusion zones.
        \item SR3: Maintain clearance from active work zones. Areas with overhead work, swinging loads, or heavy equipment are avoided unless explicitly flagged as safe.
        \item SR4: Respect unstable or unfinished surfaces. Fresh concrete, backfilled trenches, mud, loose gravel, or areas without a permanent subfloor are considered unsafe.
    \end{itemize}

\subsection{Constraint Aware Planning}

\begin{figure*}[!tb]
    \centering
    \includegraphics[width=1\textwidth]{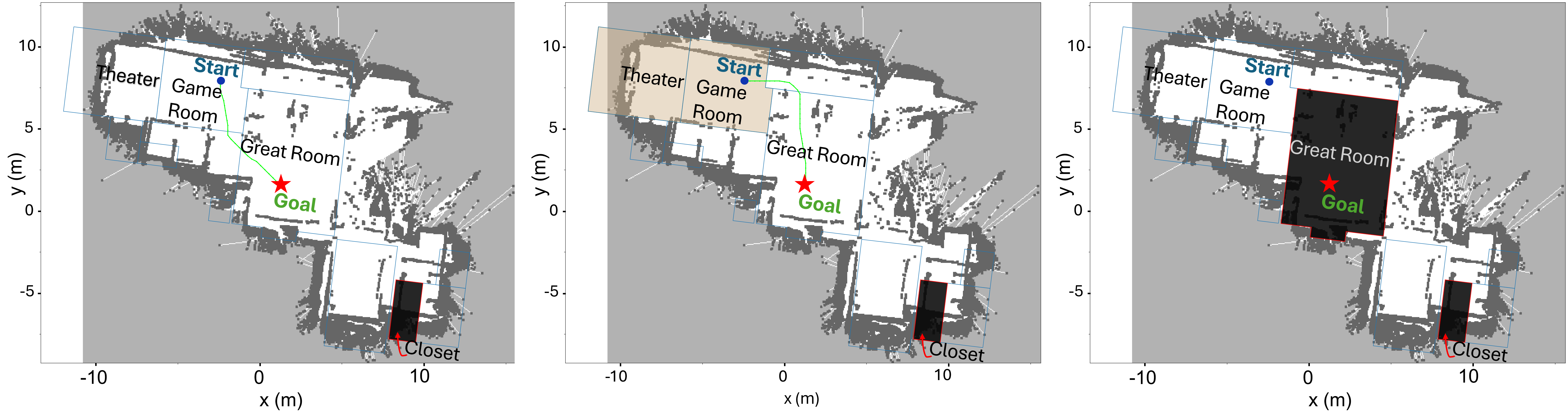}
    \caption{Schedule-aware navigation under changing construction constraints. The same navigation request is evaluated at different schedule states, demonstrating adaptation to inactive, soft, and hard work zones. Left: no active constraints affect the route, resulting in the baseline path. Center: active soft zones (tan) introduce a cost penalty. The planner exits the start region and routes the remainder of the path through unaffected rooms, minimizing traversal of the penalized area rather than continuing through it as in the unconstrained case. Right: an active hard zone (black) overlaps the goal region, causing the planner to correctly return an infeasible result. Blue circle: robot start; red star: goal; blue polygons: blueprint room boundaries; green: planned path.}
    \label{Fig 4.}
\end{figure*}

The validated constraints are incorporated directly into the planner. Hard zones are rasterized into the occupancy grid as occupied cells, making them indistinguishable from static obstacles during graph search. No feasible path can therefore intersect an active hard zone, and a task whose destination lies inside one is reported as infeasible. Soft zones remain in the traversable search space but carry additional traversal cost, encouraging rather than enforcing avoidance (Fig.~\ref{Fig 4.}).

The planner constructs a spatial cost field combining obstacle clearance
costs and constraint-dependent penalties. For each free-space cell, the
base traversal cost is $C(x) = 1 + \frac{w_0}{d(x) + \varepsilon}.$
Here, $d(x)$ is the distance (in grid cells) from cell $x$ to the nearest excluded cell (e.g., obstacle),
$w_0$ is an obstacle proximity weight, and $\varepsilon$ prevents
division by zero. We set $w_0=3.0$ (in grid cells) to bias the planner away from wall-hugging paths and $\varepsilon=10^{-3}$. Following the potential-field principle of increasing obstacle repulsion at close range~\cite{khatib1985}, we assign higher costs to lower-clearance cells within a costmap-based framework~\cite{lu2014layered}, using an inverse-clearance penalty~\cite{karakaya2026}. Cells inside active soft zones incur a cost $C_{\mathrm{soft}}(x) = \lambda\, C(x),$ where $\lambda$ is the soft-zone penalty multiplier set to 5. Soft constraints, therefore, do not prohibit traversal but bias the planner toward lower-cost alternatives. In cases where avoidance requires excessive detour or no alternative exists, the planner may traverse the region while remaining
feasible. An A* search is performed over the resulting cost field to generate the minimum-cost feasible path. Following planning, the trajectory is resampled into uniformly spaced waypoints for execution within ROS 2.

\section{Experiments and Results}
\subsection{Sites and Platforms}

The proposed framework is evaluated in two environments: (1) an office floor, and (2) an active construction site, which tests the system under realistic operating conditions. All quantitative results were computed offline using RGB, depth, and pose data collected on a Unitree Go2 during site surveys; the planning pipeline was evaluated on this recorded sensor data. From this recorded data, we evaluate multiple scenarios spanning different days and times of day, as scheduled work zones become active. 
Live deployment of the full pipeline was qualitatively demonstrated on both the Unitree Go2 quadruped and the G1 humanoid.

\subsection{Evaluation Metrics}

Navigation performance is evaluated using the following metrics:
\begin{itemize}
    \item Constraint Compliance Rate (CCR): percentage of trajectories
    remaining outside all active hard exclusion zones.
    \item Schedule-Aware Compliance Rate (SACR): percentage of feasible trajectories avoiding soft active zones, at the query time. Tasks without a feasible trajectory are counted as unsuccessful.
    \item Task Success Rate (TSR): percentage of feasible tasks completed
    successfully.
    \item Path Efficiency Ratio (PER): planned path length divided by the length of the path planned for the same task with no active schedule constraints.
    \item Zone Entry Rate (ZER): percentage of trials in which the robot enters
    a hazardous zone (i.e., one which must have been a hard constraint) mislabeled as a soft constraint.
    \item Feasible Planning Rate (FPR): percentage of requests for which the
    planner returns a feasible path under all active constraints.
\end{itemize}

\subsection{Blueprint-Based Room Resolution}

We first evaluate blueprint-to-map alignment across the three construction-site floors and the office. Alignment quality is quantified by the root mean square error (RMSE) between transformed blueprint wall points and the corresponding occupancy-map walls, with lower values indicating a better fit. Table~\ref{tab:registration} summarizes the results.

\begin{table}[htb!]
\centering
\caption{Blueprint-to-Occupancy Map Registration Accuracy}
\label{tab:registration}
\footnotesize
\renewcommand{\arraystretch}{1.15}
\begin{tabular}{lc}
\hline
\textbf{Site/Floor} &
\textbf{RMSE (m)}\\
\hline
Construction floor 0 & 0.386 \\
Construction floor 1 & 0.280 \\
Construction floor 2 & 0.366 \\
Office & 0.135 \\
\hline
\end{tabular}
\end{table}

The variation in RMSE across environments reflects differences in the quality of the robot-generated occupancy maps and the geometric structure available from the CAD drawing. Larger mapping areas may accumulate greater localization or odometry error, while differences in observable wall geometry and map completeness can also affect ICP alignment quality.

\begin{table}[!t]
\centering
\caption{Room Label Resolution Accuracy}
\label{tab:room-resolution}
\footnotesize
\renewcommand{\arraystretch}{1.15}
\setlength{\tabcolsep}{3pt}
\begin{tabular}{%
  >{\raggedright\arraybackslash}p{2.4cm}
  >{\centering\arraybackslash}p{1.4cm}
  >{\centering\arraybackslash}p{1.4cm}
  >{\centering\arraybackslash}p{1.5cm}
  >{\centering\arraybackslash}p{1.0cm}}
\hline
\textbf{Site/ Floor} &
\textbf{Room segments (HOV-SG/ CLIP)} &
\textbf{Blueprint rooms (ground truth)} &
\textbf{Labels matching blueprint} &
\textbf{Acc. (\%)} \\
\hline
Construction floor 0 & 10 & 11 & 0 & 0 \\
Construction floor 1 & 2  & 18 & 0 & 0 \\
Construction floor 2 & 12 & 21 & 0 & 0 \\
Office               & 6  & 4  & 1 & 16.7 \\
\hline
Overall              & 30 & 54 & 1 & 3.3 \\
\hline
\end{tabular}
\end{table}

\begin{table*}[!tb]
\centering
\caption{Temporal Constraint Efficiency}
\label{tab:temporal}
\footnotesize
\renewcommand{\arraystretch}{1.15}
\begin{tabularx}{\textwidth}{l
  >{\hsize=0.50\hsize\centering\arraybackslash}X
  >{\hsize=0.58\hsize\centering\arraybackslash}X
  >{\hsize=1.42\hsize\centering\arraybackslash}X
  >{\hsize=1.24\hsize\centering\arraybackslash}X
  >{\hsize=1.24\hsize\centering\arraybackslash}X
  >{\hsize=0.76\hsize\centering\arraybackslash}X
  >{\hsize=1.26\hsize\centering\arraybackslash}X}
\hline
\textbf{Site/Floor} &
\textbf{Trials} &
\textbf{Feasible} &
\textbf{Hard Violations (\%)} &
\textbf{Soft Avoided (\%)} &
\textbf{Soft Crossed (\%)} &
\textbf{SACR (\%)} &
\textbf{Mean PER} \\
\hline
Construction floor 0 & 22 & 20 & 0 & 0.0 & 100.0 & 77.3 & $1.027\pm0.064$\\
Construction floor 1 & 23 & 19 & 0 & 78.9 & 21.1 & 65.2 & $1.028\pm0.054$\\
Construction floor 2 & 23 & 18 & 0 & 0.0 & 100.0 & 65.2 & $1.000\pm0.000$\\
Office & 21 & 17 & 0 & 0.0 & 100.0 & 23.8 & $1.001\pm0.002$\\
\hline
\end{tabularx}
\end{table*}

We next compare HOV-SG's room resolution pipeline \cite{hovsg} with deterministic blueprint-based lookup across all four sites. HOV-SG partitions the mapped space into room segments using geometric heuristics, then labels each segment by matching its visual CLIP embedding against text embeddings of candidate room names (e.g., office, kitchen, bathroom) and selecting the highest-similarity category. The blueprint approach instead obtains labels directly from the CAD drawing by determining which room polygon contains the target location.

Table~\ref{tab:room-resolution} summarizes the comparison. HOV-SG produced 30 room segments while the blueprint defines 54 rooms, with substantial variation across floors. A segment was counted correct only when its assigned label matched the blueprint label for the room it overlapped. Under this criterion, CLIP-based labels matched in one of 30 segments (3.3\%), with 0\% accuracy on all three construction floors and 16.7\% in the office. Manual inspection showed that segment boundaries often diverged substantially from the corresponding architectural rooms. Because construction schedules and safety constraints are defined per room, unstable segment boundaries and unreliable labels propagate directly into incorrect constraint activation, motivating the deterministic blueprint lookup used in our approach.

\subsection{Schedule-Aware Navigation}

Next, we evaluate the system's ability to adapt navigation behavior as construction activities change over time. The same navigation request is issued at multiple times of the day as different work zones become active, and the resulting trajectories are compared using SACR and PER metrics. The experiment comprises 89 trials, one per distinct schedule timestamp swept across the four weekly look-ahead project schedules.

Table~\ref{tab:temporal} summarizes the results. Of the 89 trials, 74 were feasible after applying the active hard constraints. Across all feasible trials, the planner produced zero hard-zone violations. Every waypoint of every planned path was tested against the active hard-zone boundaries to confirm that none fell inside a restricted region. The remaining 15 trials were infeasible, as an endpoint fell inside an active hard zone in every case: 6 with the goal enclosed and 9 with the start. No trial failed from route severance between otherwise accessible endpoints. In each case the planner correctly reported infeasibility rather than returning an invalid path.

SACR is a strict measure. A trial is compliant only if the planner returned a path that intersected no active zone. It varies from 77.3\% on construction floor 0 to 23.8\% in the office, for two reasons, neither reflecting a failure to avoid zones. First, trials correctly rejected because an active hard zone blocks the start or goal do not contribute to SACR. Second, at three of the four sites, every soft-active trial involved a zone the robot could not avoid. The office illustrates both: of 21 requests, 4 were rejected for a blocked start due to a hard zone, and in 12 the active soft region occupied the only corridor between start and goal and was crossed every time. Only the 5 remaining trials, with no active zone, could contribute to SACR, giving 23.8\%. SACR is thus best read alongside the soft-avoided and soft-crossed columns, which separate zones the planner chose to cross from those it had no route around. 
Soft-constraint interaction was analyzed for trials with at least one active soft zone. The planner traded off constraint avoidance against path efficiency according to the availability of alternative routes. Construction floor 1 showed the clearest avoidance, routing around active soft regions in 78.9\% of soft-active trials and entering them only when the alternative was less favorable. Construction floors 0 and 2 and the office showed 100\% crossings, but these reflect geometry rather than ineffective avoidance: each was an unavoidable interaction. Even then, the graded cost planner still shapes the path. In Fig.~\ref{Fig 4.}, the planner exits the penalized region early and completes the route through unaffected rooms, accepting additional distance for reduced traversal of the soft zone.

Path efficiency remained consistent across environments, with mean PER near 1.0. The largest variations occurred on construction floors 0 and 1. On floor 1, the planner selected longer routes to avoid active soft regions in most soft-active trials; on floor 0, the variation came from clearance-constrained detours around narrow passages rather than constraint avoidance.

\subsection{Safety Rule Validation}

This experiment evaluates the LLM-based safety validation module under incorrect operator annotations. For each rule we built one hazardous and one non-hazardous test case, eight in total. In each hazardous case a work zone is deliberately labeled soft while its notes accurately describe the hazard; the task is then executed with validation disabled and enabled. With validation enabled, the model escalates the zone to a hard constraint according to the codified safety rules, and the planner reroutes around the hazard. ZER is measured across multiple tasks to quantify the protection added by the validation layer.
Table~\ref{tab:escalation} summarizes the eight escalation scenarios. All four hazardous cases were correctly escalated to hard constraints and all four non-hazardous cases remained soft, giving complete agreement with the expected outcomes. Enabling validation reduced ZER from 100\% to 0.0\% for the hazardous cases while leaving the non-hazardous cases unchanged. To characterize stability, each case was scored eight times (64 calls). Seven cases showed zero score variance; the remaining case (SR2 hazardous) had a standard deviation of 0.024. No call crossed the escalation threshold in the wrong direction, and all escalation decisions were identical across repetitions, indicating that the validation module's behavior is stable.

\begin{table}[t]
\centering
\caption{Safety Escalation Outcomes}
\label{tab:escalation}
\footnotesize
\renewcommand{\arraystretch}{1.15}
\resizebox{\columnwidth}{!}{%
\begin{tabular}{lccc}
\hline
\textbf{Case} &
\textbf{Score (mean, $N{=}8$)} &
\textbf{Escalated} &
\textbf{Expected} \\
\hline
SR1 Hazardous & 1.00 & Yes & Yes \\
SR1 Non-hazardous & 0.00 & No  & No  \\
SR2 Hazardous & 0.92 & Yes & Yes \\
SR2 Non-hazardous & 0.00 & No  & No  \\
SR3 Hazardous & 0.95 & Yes & Yes \\
SR3 Non-hazardous & 0.00 & No  & No  \\
SR4 Hazardous & 1.00 & Yes & Yes \\
SR4 Non-hazardous & 0.00 & No  & No  \\
\hline
\end{tabular}%
}
\end{table}

\subsection{Ablation Study}

To evaluate the contribution of each system component, we perform ablation studies by removing (1) construction schedules, (2) blueprint-based room resolution, and (3) LLM-based safety rule validation. Each ablation is compared against the complete system to quantify its impact on navigation performance and safety.

In addition to the component ablations, we compare against HOV-SG \cite{hovsg} used directly as a navigation system, in which targets are grounded and localized through its scene graph and CLIP-based retrieval alone. As no prior method targets blueprint- and schedule-aware navigation without a BIM model, no directly comparable external system exists; existing BIM-based navigation approaches \cite{robotics10010002} \cite{gopee2023ifc} require IFC models that are unavailable for our test sites. HOV-SG therefore serves as the strongest available semantic-navigation baseline and represents the perception layer our framework builds upon, evaluated without construction-specific grounding, scheduling, or safety validation. The complete system, the three ablations, and the HOV-SG baseline, are all evaluated on the same 54 navigation requests, so differences reflect only the method under test.

Table~\ref{tab:ablation} summarizes the ablation results. The complete CORNAV pipeline achieves 100\% CCR and 72.2\% TSR at a mean PER of 0.988. It returns a feasible path for 46.3\% of requests once all active constraints are enforced, and every path it returns satisfies all mandatory safety requirements. %demonstrating that the integrated system correctly prioritizes safety over task completion. 

\begin{table}[htb!]
\centering
\caption{Comparison of Navigation Planning Methods}
\label{tab:ablation}
\footnotesize
\renewcommand{\arraystretch}{1.15}
\setlength{\tabcolsep}{3pt}
\resizebox{\columnwidth}{!}{
\begin{tabular}{lccccc}
\hline
\textbf{Method} &
\textbf{CCR (\%)} &
\textbf{TSR$^{\dagger}$ (\%)} &
\textbf{FPR (\%)} &
\textbf{SACR (\%)} &
\textbf{PER}$^{\ddagger}$ \\
\hline
CORNAV \textbf{(ours)}          & 100.0 & 72.2 & 46.3 & 29.6 & 0.988 \\
w/o Blueprint               & 98.1 & 13.0  & 40.7 & 25.9 & 0.984 \\
w/o Schedules               & 53.7  & 72.2 & 40.7 & 24.1 & 1.000$^{*}$ \\
w/o Safety Validation       & 83.3 & 72.2 & 46.3 & 29.6 & 0.991 \\
HOV-SG Baseline\cite{hovsg}             & 42.6  & 7.4   & 31.5 & 20.4 & 1.398 \\
\hline
\end{tabular}}

\vspace{2pt}
\begin{minipage}{\columnwidth}
\scriptsize
$^{\ddagger}$PER uses clearance-weighted cost, not raw distance, and constraints in one region can shift the globally cost-optimal route elsewhere. The constrained path can therefore be geometrically shorter than its reference, so PER values marginally below 1.0 do not indicate error.\\
$^{\dagger}$CORNAV, w/o Schedules, and w/o Safety Validation share the same object-grounding stage; the schedule and safety-validation flags affect only planning, after the target is identified. Their TSR values are therefore identical by construction.\\
$^{*}$Reference condition for PER.
\end{minipage}
\end{table}

\begin{table}[t]
\centering
\caption{Task Success Rate by Object Source}
\label{tab:object-source}
\footnotesize
\renewcommand{\arraystretch}{1.15}
\begin{tabular}{lcc}
\hline
\textbf{Method} &
\textbf{Permanent Object} &
\textbf{CLIP} \\
\hline
CORNAV \textbf{(ours)} & 100.0 & 63.4 \\
w/o Blueprint & 0.0 & 17.1 \\
w/o Schedules & 100.0 & 63.4 \\
w/o Safety Validation & 100.0 & 63.4 \\
HOV-SG Baseline \cite{hovsg}   & 0.0 & 9.8 \\
\hline
\end{tabular}
\end{table}

CORNAV and the variant without safety validation both reach 29.6\% on SACR, the highest of all variants. This tie does not mean they are equally safe. CORNAV reaches 29.6\% by declining tasks that would cross a hazardous zone, while the ablation reaches it by attempting more tasks and failing the hazardous ones, so its extra attempts and extra failures cancel out. The gap in CCR (100\% vs 83.3\%) reflects this difference.

Removing blueprint-based grounding produces the largest reduction in task performance. TSR decreases from 72.2\% to 13.0\%, with retrieval of permanent architectural fixtures failing completely (0.0\% TSR) and CLIP-based object retrieval dropping to 17.1\% (refer to  Table~\ref{tab:object-source}). 
Deterministic blueprint-based room grounding is thus essential for reliable object localization, especially for permanent building features that cannot be robustly identified through semantic similarity alone. 

Removing schedule awareness produces the largest compliance drop of any component ablation, reducing CCR from 100.0\% to 53.7\%. 
Without temporal information, the planner cannot tell which hard zones are active and routes through them, so it still returns feasible paths at the blueprint ablation's rate (40.7\% FPR) while its compliance collapses. FPR and CCR are thus distinct axes. A planner can find paths just as often while being far less safe.

Disabling safety validation module leaves CCR at 83.3\% and FPR at 46.3\%. Both CORNAV and this ablation find a feasible path for the same share of requests, so their FPR is identical. The difference between the two arms lies in validation, not path planning. CORNAV recognizes that some of these paths cross promoted hazard zones and refuses them, while the ablation, lacking that check, executes them. Executing these hazardous paths lowers its CCR to 83.3\% while leaving its planning rate unchanged.

The HOV-SG baseline exhibits the weakest overall performance, achieving 7.4\% TSR and 42.6\% CCR while producing feasible paths for 31.5\% of navigation requests. Operating without blueprint grounding, schedule awareness, or construction-specific safety reasoning, it frequently navigates toward incorrect targets while traversing active work zones. These results demonstrate that semantic scene understanding alone is insufficient for reliable navigation on active construction sites, and that integrating architectural, temporal, and safety information substantially improves both navigation accuracy and operational safety.

We also performed 12 experiments with the Go2 quadruped and the G1 humanoid for various missions using the full CORNAV pipeline. In all cases, the robots were able to take the schedule into account and reach the destination. When a path was infeasible under the constraints set by the project schedule, CORNAV correctly identified the infeasibility and the robots did not move. The videos for select task queries are available on our website.

\section{Conclusion}

This paper presented CORNAV, a blueprint-grounded and schedule-aware navigation framework for autonomous robots in dynamic construction environments. By combining architectural blueprints, hierarchical open-vocabulary 3D scene graphs, project schedules, and LLM-based safety validation in a unified planning pipeline, the system reasons not only about where objects are located, but when areas are safe to access. Experiments in an indoor office and active construction settings showed that blueprint-based grounding substantially improves object localization, schedule-aware planning prevents navigation through active work zones, and safety validation guards against incorrectly labeled hazards. Together, these capabilities enable reliable navigation under construction-specific constraints. The presented pipeline provides a foundation for future work, including multi-robot coordination on a shared construction site, and tighter coupling between task planning and mobile manipulation for long-horizon tasks.

\section{Acknowledgments}
The authors thank Genova Construction and Shane Stanfield for providing the Unitree G1 humanoid robot and access to the construction site.

\bibliographystyle{IEEEtran}
\bibliography{cite}

\end{document}